\documentclass[11pt]{article}

\usepackage{geometry}
\usepackage{setspace}
\usepackage{caption} 
\usepackage{graphicx}
\usepackage{url}

\usepackage{setspace}
\usepackage{caption} 
\usepackage{subcaption}
\usepackage{pifont}
\usepackage{booktabs}
\usepackage{tcolorbox}
\usepackage{xcolor}

\usepackage{mathptmx}
\usepackage[T1]{fontenc}

\tcbuselibrary{theorems}

\newtcbtheorem[number within=section]{customdef}{Definition}%
{colback=gray!5,colframe=gray!35!black,fonttitle=\bfseries}{th}
\usepackage{cleveref}

\usepackage{natbib}
\setcitestyle{aysep={}}

\title{Data-Centric Artificial Intelligence} 

\author
{Johannes Jakubik$^{1,\ast}$, Michael Vössing$^{1}$, Niklas Kühl$^{2}$, Jannis Walk$^{1}$, Gerhard Satzger$^{1}$\\
\\
\normalsize{$^{1}$Karlsruhe Institute of Technology}\\ 
\normalsize{Kaiserstraße 12, 76131 Karlsruhe, Germany}\\[5pt]
\normalsize{$^{2}$University of Bayreuth}\\ 
\normalsize{Universitätsstraße 30, 95447 Bayreuth, Germany}\\[5pt]
\normalsize{$^{\ast}$Corresponding author: Johannes Jakubik, E-mail: johannes.jakubik@kit.edu}
}

\date{}

\begin{document} 

\baselineskip18pt


\maketitle




\begin{center}
    \textbf{Accepted for Publication at \textit{Business \& Information Systems Engineering}}
\end{center}

\section{Introduction}
\label{sec:intro}

{Over the past decades, researchers and practitioners in artificial intelligence (AI) have focused on improving ML models in AI-based systems (model-centric AI paradigm). However, the provision and selection of suitable data also impact model effectiveness (e.g., performance) and efficiency (e.g., costs for labeling or for training computation). Despite a long history of research on data \citep{Legner.2020,Otto2011,zhang2019discovering}, the impact of data quantity and quality on AI-based systems is still often overlooked in both AI research \citep{Parmiggiani2022} and AI practice \citep{Sambasivan2021}. 
Propagated by Andrew Ng and promoted in a series of workshops \citep{Ng2021NIPS,Ng2022competition}, data-centric AI emphasizes the development and application of methods, tools, and best practices for systematically designing datasets and for engineering data quality and quantity to improve the performance of AI-based systems \citep{Strickland2022}.} {In particular, the new paradigm is not calling for simply acquiring more data but more appropriate data.} While many facets of data-centric AI have previously been studied independently, this paradigm unites researchers from different fields (e.g., machine learning and data science, data engineering, and information systems) with the goal of improving machine learning approaches in real-world settings. This has far-reaching implications for the way AI-based systems are developed.

The objective of this article is to introduce practitioners and researchers {in Business and Information System Engineering to data-centric AI as a complementary and mutually beneficial paradigm to model-centric AI}. We define relevant terms, introduce key characteristics to contrast both paradigms, and introduce a framework for data-centric AI. We distinguish data-centric AI from related concepts and, in particular, discuss potential contributions of and implications for the BISE community.

\section{{Model-centric and Data-centric AI}}

In previous years, research on ML has mainly focused on {the development of model types, architectures, and the definition of suitable hyperparameters} to improve performance. For example, the ML community often benchmarks different ML approaches based on fixed datasets---both in practical competitions \citep{Kaggle} as well as in academic research \citep[e.g., ][]{Ronneberger2015}. {Utilizing publicly available benchmark datasets allows for valuable and scientific sound comparisons across approaches and has facilitated a significant acceleration in the performance of ML models. In addition, these benchmark datasets can be employed to ensure the reproducability of proposed models. Overall, this led to an increasing maturity of model types, architectures, and hyperparameter selection.}
\begin{customdef*}{Model-centric Artificial Intelligence}{}
    {\textit{Model-centric artificial intelligence} is the paradigm {focusing on the choice of the suitable model type, architecture, and hyperparameters from a wide range of possibilities for building effective and efficient AI-based systems}.
    }
\end{customdef*}
{However, in recent years, this strategy (i.e., solely optimizing models) has plateaued for many datasets with regard to the model performance. Similarly, with regard to real-world datasets, a focus on improving (complex) models does not necessarily lead to significant performance increases \citep[e.g., ][]{Baesens.2021}}. Furthermore, practitioners often want to use ML to solve unique problems for which neither public datasets nor suitable pre-trained models are available.
%
For this reason, the focus of practitioners and researchers has gradually been shifting towards data, the second, somewhat neglected ingredient for the development of AI-based systems. In particular, researchers and practitioners recognize the need for more systematic data work as a means to improve the data used to train ML models. 
In fact, data is a crucial lever for an ML model to generate knowledge \citep{Groger.2021}. Consequently, data quantity (e.g.,  the number of instances) and data quality (e.g., data relevance and label quality) largely influence the performance of AI-based systems \citep{Gudivada.2017}. Data-centric artificial intelligence (data-centric AI) represents a paradigm that reflects this. 

\begin{customdef*}{Data-centric Artificial Intelligence}{}
    \textit{Data-centric artificial intelligence} is the paradigm emphasizing that the systematic design and engineering of data are essential for building effective and efficient AI-based systems.
\end{customdef*}

\noindent Data-centric AI differs from model-centric AI in terms of the general focus, the importance of domain knowledge, and the understanding of data quality: 
\begin{itemize}
    \item \textbf{Focus}: {Data-centric AI generally holds the ML model fixed instead of the dataset.} Performance increases are achieved by improving the quality and quantity of the data given a fixed model.
    \item \textbf{Data Work and Domain Knowledge}: Domain-specific data work is an integral component of data-centric AI. Data work is supplemented by the development of methods and semi-automated tools to accelerate the development of successful AI-based systems.
    \item \textbf{{Perspective on Data Quality}}: 
    Data-centric AI generates performance improvements based on more appropriate data. This implies that changes in ML model performance metrics also indicate the effectiveness of adjustments in the data. This results in a novel perspective on data quality that can be approximated by changes in metrics from the field of machine learning.
\end{itemize}

\begin{figure}[ht]
    \centering
    \includegraphics[width=0.8\linewidth]{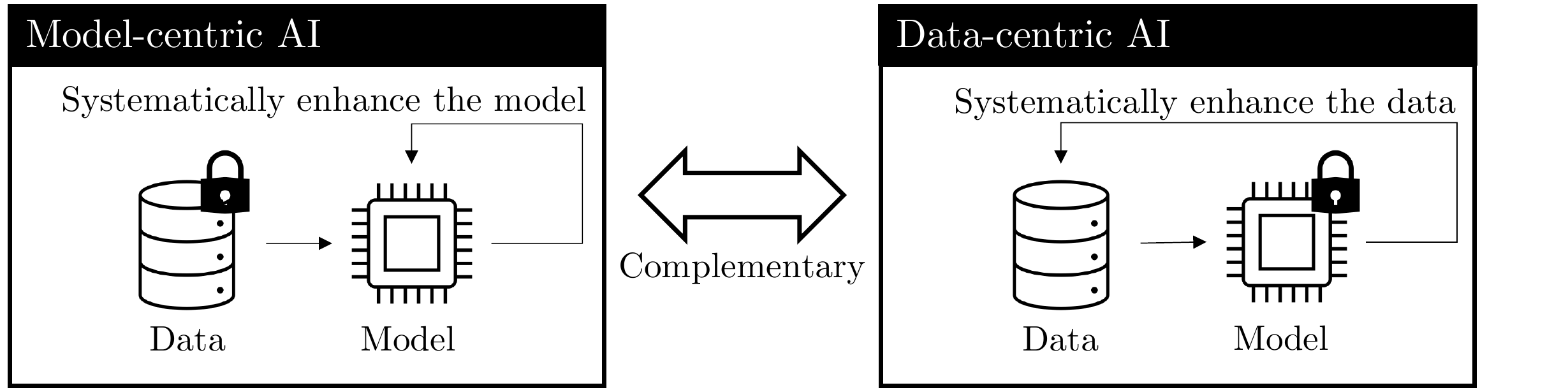}
    \caption{{Data-centric AI as an emerging, complementary paradigm for the development of AI-based systems.}}
    \label{fig:model-vs-data}
\end{figure}

Despite these differences between model-centric and data-centric AI, the two paradigms are inherently complementary, as the development of AI-based systems should ultimately incorporate both paradigms. A high-level overview depicting this relationship is displayed in \Cref{fig:model-vs-data}. 

While the data-centric paradigm emerges from the ML community---and most academic endeavors dealing with it do focus on machine learning---, {the term ``data-centric AI'' has also intruded the computer science and BISE communities. However, in fact, data-centric \textit{machine learning} might have been a more appropriate term {\citep{kuhl2020machine}}}. ML research generally focuses on designing methods that leverage data to increase the performance on a range of tasks (i.e., learn) with computational resources \citep{alpaydin2021machine}. Artificial intelligence, in contrast, includes ML but also comprises a broader set of methods---e.g., logical programming or probabilistic methods---that allow an agent to interact with its environment. 

\section{{Dimensions of Data-centric AI}}

\begin{figure}
    \centering
    \includegraphics[width=\linewidth]{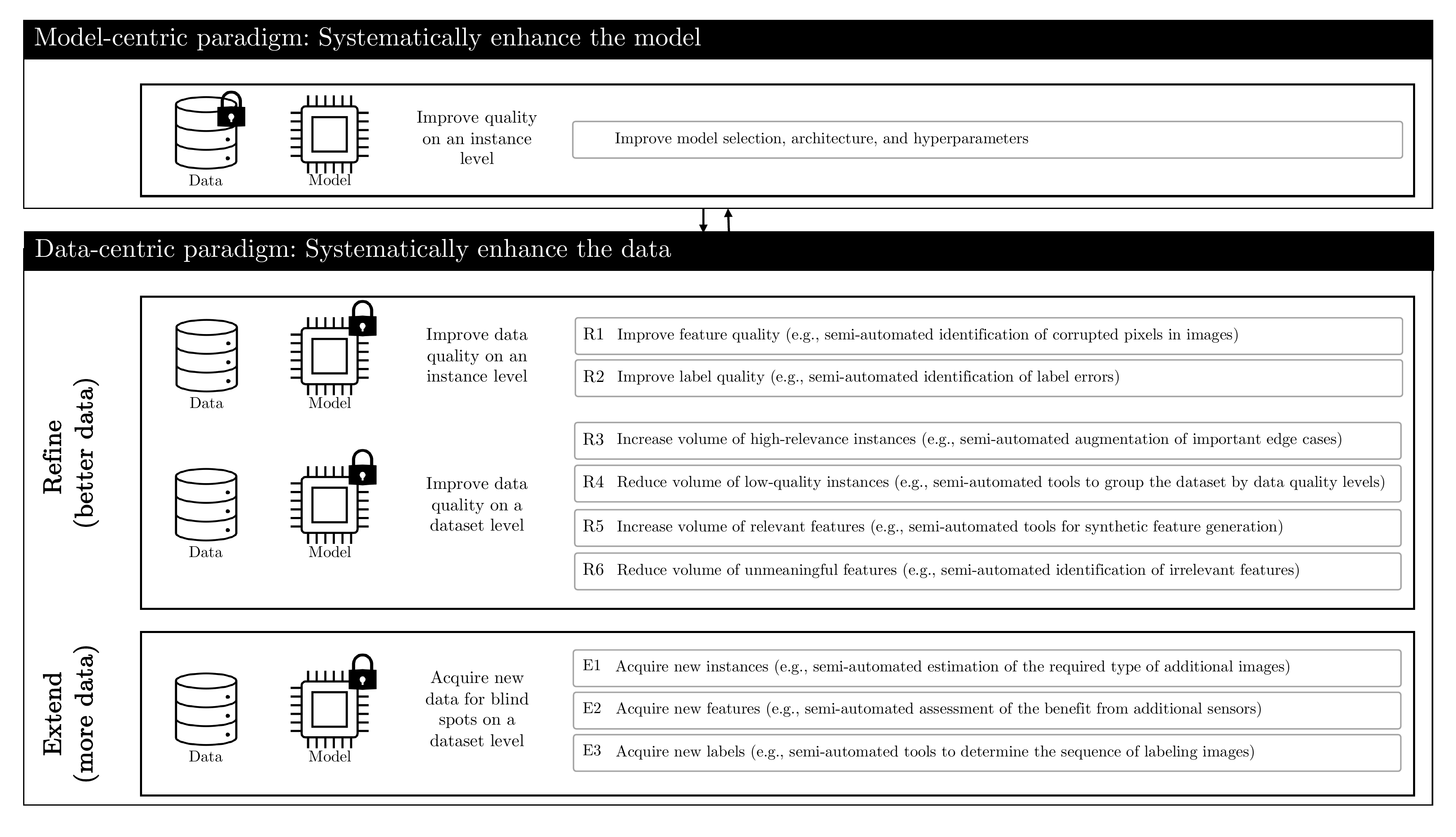}
    \caption{{Framework for the systematic design and engineering of data for data-centric AI.}\protect\footnotemark}
    \label{fig:data-centric-ai}
\end{figure}

{The framework for data-centric AI in \Cref{fig:data-centric-ai} illustrates the different dimensions for the systematic design and engineering of data.}
While data-centric AI is also applicable to unsupervised \citep{amrani2021model} or reinforcement \citep{lin2022roboflow} learning, this summary focuses on supervised ML as the most prevalent real-world application of ML \citep{jordan2015machine}. Overall, we identify two major dimensions for data-centric AI---that is, the \emph{refinement} of existing data (i.e., ``better data'') and the \emph{extension} of this data by acquiring additional data (i.e., ``more data''). 

The \textit{refinement} of data refers to systematically improving the quality of existing data---measured with performance metrics of ML models. First, enhancing the quality of individual instances can be achieved by improving the {quality of features or target labels} (R1, R2). Regarding the representativeness of the data, the quality of data can be enhanced by increasing the number of high-relevance instances that strongly influence the learning process of ML models (R3). Such underrepresented but relevant instances need to be particularly taken into account for augmentations. {Moreover, low-quality instances with, for example, incorrect labels or inaccurate feature values need to be identified and removed from the data (e.g.,  the semi-automatic identification of label errors in R4; see \citeauthor{northcutt2021pervasive}, \citeyear{northcutt2021pervasive}). Thus, it is essential for data-centric AI to build semi-automated tools to better differentiate outliers (that should be removed from the dataset) from edge cases (that should be augmented to enhance the representativeness of the data)}. On a feature level, data refinement means increasing the volume of relevant features while excluding unmeaningful or even unfair {ones}. While these actions to refine datasets have partly been leveraged in the past, this work has mostly been {performed} manually.

The \textit{extension} of data refers to systematically acquiring additional data for ``blind spots'' in the dataset.
Extending the data becomes necessary when the existing data does not allow to sufficiently address the business problem. Additional data may help to develop an accurate ML model. There are two {major motivations} for extending the data: First, extending data helps to achieve an initial ML model performance that meets the requirements of the business problem. Second, extending the dataset helps to respond to shifts in the data distribution to maintain this performance over time. Hence, acquiring new data is crucial for both achieving high performance of the model and maintaining the performance of the AI-based system. Overall, we identify three dimensions along which data can be extended: First, new {instances} may be acquired, whereby each instance represents an observation (E1). Second, additional, new {features} for each of the instances may be collected, e.g.,  by employing additional sensors (E2). Finally, collecting additional data may also refer to the retrieval of {target labels} for existing or new unlabeled instances (E3). Overall, extending the data by acquiring new features, observations, or target labels (E1--E3) primarily impacts the quality of data based on increasing data {quantity} (more data). In contrast, refining existing data targets the improvement of {data quality based on operation with existing data (better data)}.
Overall, both dimensions of the framework are heavily influenced by major IS topics{,} including, for example, data governance, data management, and AI governance. We discuss the relation between these topics and data-centric AI in the last section and use our framework to better link the importance of these topics to data-centric AI.

\footnotetext{{Note that we illustrate a high-level representation of model-centric AI only. For more details on model-centric AI, we refer to \citet{alpaydin2021machine}.}}

For widespread adoption of data-centric AI in research and practice, methods and appropriate tools are required. While some methods and {corresponding} measures already exist, e.g., for the identification of special instances in the data, there is a particular void of methods to systematically design and engineer the data. This includes, among others, methods for data versioning \citep[e.g., ][]{wandb}, methods that support the labeling process in terms of efficiency and performance \citep[e.g.,  R2, E3; see][]{fiedler2018imagetagger}, methods for data exploration {and visualization} \citep[e.g.,  R1, R2; see][]{mcinnes2018umap}, and methods to identify special data instances \cite[e.g.,  R1--R6; see][]{northcutt2021pervasive}. {We provide a summary on data-centric tools, including examples of commercial applications, in Table~\ref{tab:tools}, where meta tools refer to tools that are of general importance for data-centric AI (i.e., across the dimensions of the data-centric AI framework).} {In general, methods from the field of transfer learning support the development of data-efficient AI-based systems across the dimensions of data-centric AI (R1--R6 and E1--E3) as pretrained models require a reduced amount of high-quality data.} From a system's perspective, methods for semi-automated data exploration are particularly important, as such methods may eventually contribute to an enhanced data understanding, which is essential for improving data quality on an instance level (R1, R2), on a dataset level (R3--R6), and for extending the data (E1--E3). {Overall, all required methods need to be supported by corresponding tools.} 

\begin{table}[ht]
    \centering
    \begin{tabular}{l|l|cccccc|ccc}
        \toprule
         Tools purpose & Commercial examples & R1 & R2 & R3 & R4 & R5 & R6 & E1 & E2 & E3   \\\midrule
         Label error identification & cleanlab.ai & & \ding{108} & & & &  & & &\\
         Labeling support & prodi.gy & & & & & &  & & & \ding{108} \\
         Synthetic data generation &  gretel.ai & & & \ding{108} & & \ding{108} &  & \ding{108} & \ding{108} &  \\
         Anomaly detection &  Microsoft Azure & \ding{108} & & & \ding{108} & & \ding{108} & & &\\
         Data gathering efficiency & --- & & & & & &  & \ding{108} & \ding{108} &\\
         Edge case identification & iMerit edge case &  & & \ding{108} & & & & \ding{108} & &\\
         Visual data exploration & tableau.com & \ding{108} & \ding{108} & \ding{108} & \ding{108} & \ding{108} &\ding{108}  & & &\\
         ... &  &  & & &\\
         \midrule
         Meta tools &  & & & & & &  & & & \\
         \midrule
         Data versioning &  wandb.ai  & \ding{108} & \ding{108} & \ding{108} & \ding{108} & \ding{108} & \ding{108} & \ding{108} & \ding{108} & \ding{108}\\
         Personalized data work & --- & \ding{108} & \ding{108} & \ding{108} & \ding{108} & \ding{108} & \ding{108} & \ding{108} & \ding{108} & \ding{108}\\
         Federated data work  & lifebit.ai & \ding{108} & \ding{108} & \ding{108} & \ding{108} & \ding{108} & \ding{108} & \ding{108} & \ding{108} & \ding{108}\\
         Data verification &  Google Cloud  & \ding{108} & \ding{108} & \ding{108} & \ding{108} & \ding{108} & \ding{108} & \ding{108} & \ding{108} & \ding{108}\\
         Data quality measurement &  precisely.com  & \ding{108} & \ding{108} & \ding{108} & \ding{108} & \ding{108} & \ding{108} & \ding{108} & \ding{108} & \ding{108}\\
         ... &  &  & & &\\
         \bottomrule
    \end{tabular}
    \caption{{Examples of data-centric tools and meta tools categorized by the dimensions of the framework for data-centric AI.}}
    \label{tab:tools}
\end{table}



\section{{Delimitations of Data-centric AI from Related Concepts}} 

The data-centric AI paradigm relates to a number of concepts that have been studied in the BISE community over the past several decades, in particular Big Data, MLOps, and data-driven methods. 
In the following, we delineate data-centric AI from these closely related concepts.

The paradigms of \textit{big data} and data-centric AI both focus on gathering more data to improve analytics and predictive tools. While the two fields are closely related, significant differences between the two exist: Big data generally refers to the collection, storage, and processing of large amounts of data (e.g., E1). {However, there is less focus on what kind of data is stored \citep{chen2012business}}. The general assumption is that more data is always better. 
In contrast, data-centric AI aims to improve the performance of AI systems by \textit{systematically} acquiring more but also better data {or even by removing deficient or irrelevant data (R1--R6). This is especially relevant in specialized domains lacking the option to collect large amounts of data}. ML models are particularly sensitive to noise in the data when data volume is small \citep{Baesens.2021}. In these cases, systematically designing and engineering datasets is crucial for the adoption and usage of AI. Moreover, data-centric AI includes additional operations on the data, such as the extension of data based on data collection in new contexts.

\textit{Machine Learning Operations (MLOps)} is another research field closely linked with the data-centric AI paradigm. MLOps is concerned with putting AI projects into production and avoiding a multitude of pitfalls associated with this process. 
To address this gap, MLOps \citep{renggli2021data}---oftentimes used interchangeably with Artificial Intelligence Operations (AIOps)---is required.
MLOps is an engineering practice dealing with the application of tools, frameworks, and best practices to increase the number of AI projects that are brought to production. 
While data does play an important role, a major part of the MLOps paradigm focuses on engineering principles like continuous development, orchestration, monitoring, reproducibility, and versioning \citep{renggli2021data}. However, so far, there is very little focus on monitoring and versioning the datasets that would be adapted by methods from the field of data-centric AI (both in R1--R6 and E1--E3). Tracking different versions of data and the corresponding impact on ML models is essential to efficiently progress towards better data to increase the performance of ML models. Therefore, tools, frameworks, and best practices are required to facilitate data work and make modifications to the data an iterative part of AI projects instead of only preprocessing the data initially while iterating the search for the optimal model.

Finally, we discuss the difference between data-centric AI and \textit{data-driven} methods due to the similarity in the terminology. Data-driven methods focus on processing data into information {in order to present the derived information to decision-makers. Model-driven methods instead focus on mathematical models like optimization or simulation models. ML models are typically considered both data-driven and  model-driven since mathematical models are fed with a large amount of data \citep[e.g., ][]{Turban2007}.} 
The distinction between model-centric and data-centric is on a lower level of abstraction{, though---}it differentiates two paradigms for developing an ML model. This means that the type of method to generate information from data (data-driven vs. model-driven) does not necessarily determine the paradigm to develop the underlying model (data-centric vs. model-centric).

\section{Implications for BISE Research}

\begin{figure}[ht]
    \centering
    \includegraphics[width=0.9\linewidth]{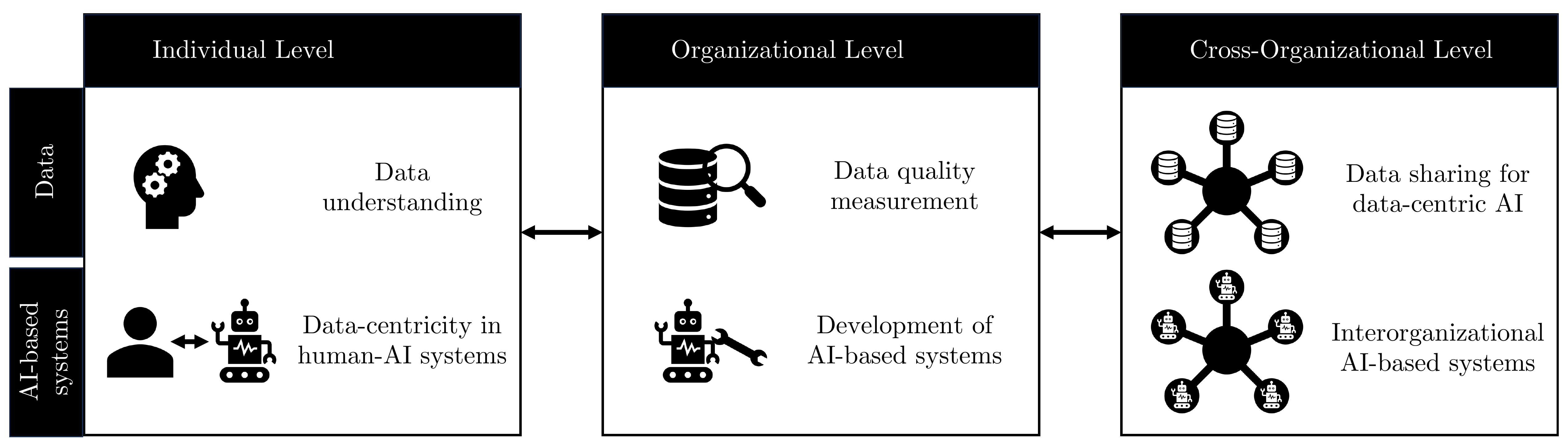}
    \caption{{Proposed areas of BISE research for the advancement of data-centric AI.}}
    \label{fig:implications}
\end{figure}

{Until now, data-centric AI has largely been explored by researchers from the field of computer science. 
However, data-centric AI has the potential to fundamentally improve AI-based systems by complementing the paradigm of model-centric AI and thereby offering a more holistic development of AI-based systems. While data-centric AI promises to support the BISE community in the design of more effective information systems, BISE researchers are also well-positioned to advance data-centric AI. {Figure~\ref{fig:implications} provides proposed areas with respect to BISE research on an individual, organizational, and cross-organizational level that are to be detailed in the following section. 
We group the proposed areas into advancements for dealing with data as such and with their incorporation within AI-based systems.}
}


\newpage
{\subsection{Individual Level}}

{
{Data-centric AI emphasizes leveraging high-quality datasets, which frequently requires to accurately select a relevant, high-quality subset of large-scale datasets. This demonstrates the significance of a nuanced data understanding for data-centric AI. Recent semi-automated methods and tools propose to generate metadata in order to improve the understanding of high-dimensional datasets \citep{holstein2023sanitizing}. With a unified set of metadata, such approaches can not only foster an understanding \textit{within} high-dimensional datasets but also comparisons \textit{across} datasets.}
Advancing data understanding relies on research in information systems, which explores data visualization and interpretability methods to help researchers and practitioners gain insights into complex data patterns \citep{toreini2022designing}. Interactive dashboards can enhance data understanding over time by allowing users to explore and analyze data from different perspectives, providing real-time insights (e.g., R1--R6). 
Building interactive data exploration tools enables users to engage with the data, facilitating ad-hoc analyses and supporting data profiling and data quality assessment. 
Additionally, information system research focuses on developing data profiling techniques that automatically extract statistical summaries, data distributions, and potential data quality issues \citep{abedjan2022data}, which help identifying missing values, outliers, and inconsistencies (R4) that may impact data-centric AI development.
{Understanding data in context, particularly within specific domains or industries, further enables data-centric AI solutions to address real-world challenges, while integrating domain-specific knowledge.}


{Data-centric AI emphasizes the importance of utilizing domain knowledge to refine and extend data towards increasing performances of AI-based systems. Thus, data-centric AI embraces both social and technical aspects by definition;} and the underlying data work for data-centric AI, {including annotation, curation, and preprocessing,} is inherently human-centered \citep{jarrahi2022principles}. 
This is, among other examples, reflected in the utilization of semi-automated tools to improve the data work of human annotators and domain experts. As a consequence, data-centric AI requires one or multiple human(s) in the loop of the AI to guarantee access to domain knowledge. 
Therefore, the efficiency of data work is mainly driven by efficient and sustainable interactions between AI and domain experts in human-in-the-loop systems (e.g., by balancing the trade-off between the value of additional data for the AI and the time and cost of data work). 
Efficient interaction is especially important as the number of manual reviews to refine data is typically constrained due to the limited availability of human experts (e.g.,  physicians; see \citeauthor{holzinger2016interactive}, \citeyear{holzinger2016interactive}). 
For example, instead of asking physicians to select and adjust incorrect labels of images depicting specific diseases, semi-automated tools are required that preselect potentially mislabeled images (e.g.,  as part of R2), {potentially important edge cases (e.g., as part of R3), or the most informative data instances for future labeling (e.g., E3, see also active learning \citep{hemmer2022deal}).} A physician can then review this subset in a fraction of the time. The design of semi-automated tools to facilitate data work requires social and technical considerations, opening various avenues for BISE research (e.g.,  personalized tools for data work, federated data work, etc.).
{The BISE community has a long history of analyzing the behavior and interactions of humans and AI, which now includes an additional facet in terms of downstream implications of human-in-the-loop systems for the improvement of data work as part of data-centric AI.}

{\subsection{Organizational Level}}




{Monitoring data quality is a critical organizational task in the context of data-centric AI \citep{schneider2023artificial}.
Recent research has demonstrated that the performance of ML models is specifically affected by incomplete data, as well as low feature and low label accuracy \citep{budach2022effects}. Monitoring the effect of data completeness, feature accuracy, and label accuracy on ML models during the enhancement of data quality in real-world datasets is essential to better understand promising ways of data enhancement \citep{aramburu2023data}. 
Ensuring appropriate data quality further requires data verification and validation, especially when dealing with real-time or sensor-generated data \citep{whang2023data,abbasi2016big}. Thus, there is a necessity for methods and tools to continuously verify and validate data (e.g., R1, R2) and provide feedback to data providers, enabling them to improve data quality. 
The shift towards the data-centric paradigm further changes the approach to measuring data quality, emphasizing continuous monitoring throughout the iterative data work process \citep{Sambasivan2021}. Quantifying data quality using ML model performance facilitates the assessment of the impact of data modifications on AI system performance across this process. Data-centric AI will benefit from a refinement in the understanding of data quality for AI-based systems and from the development of tools to continuously measure data quality in AI projects. 
Additionally, guidance from BISE researchers is needed to investigate innovative ways to seamlessly integrate diverse, heterogeneous data sources (e.g., E2), overcoming the limitation of current data integration approaches and enabling comprehensive data-centric AI applications \citep{grover2018creating}. 
{Through these efforts, BISE research can help to empower data-centric AI to harness high-quality data, improving the performance of AI-based systems across domains and industries.}


\begin{figure}[ht]
    \centering
    \includegraphics[width=0.9\linewidth]{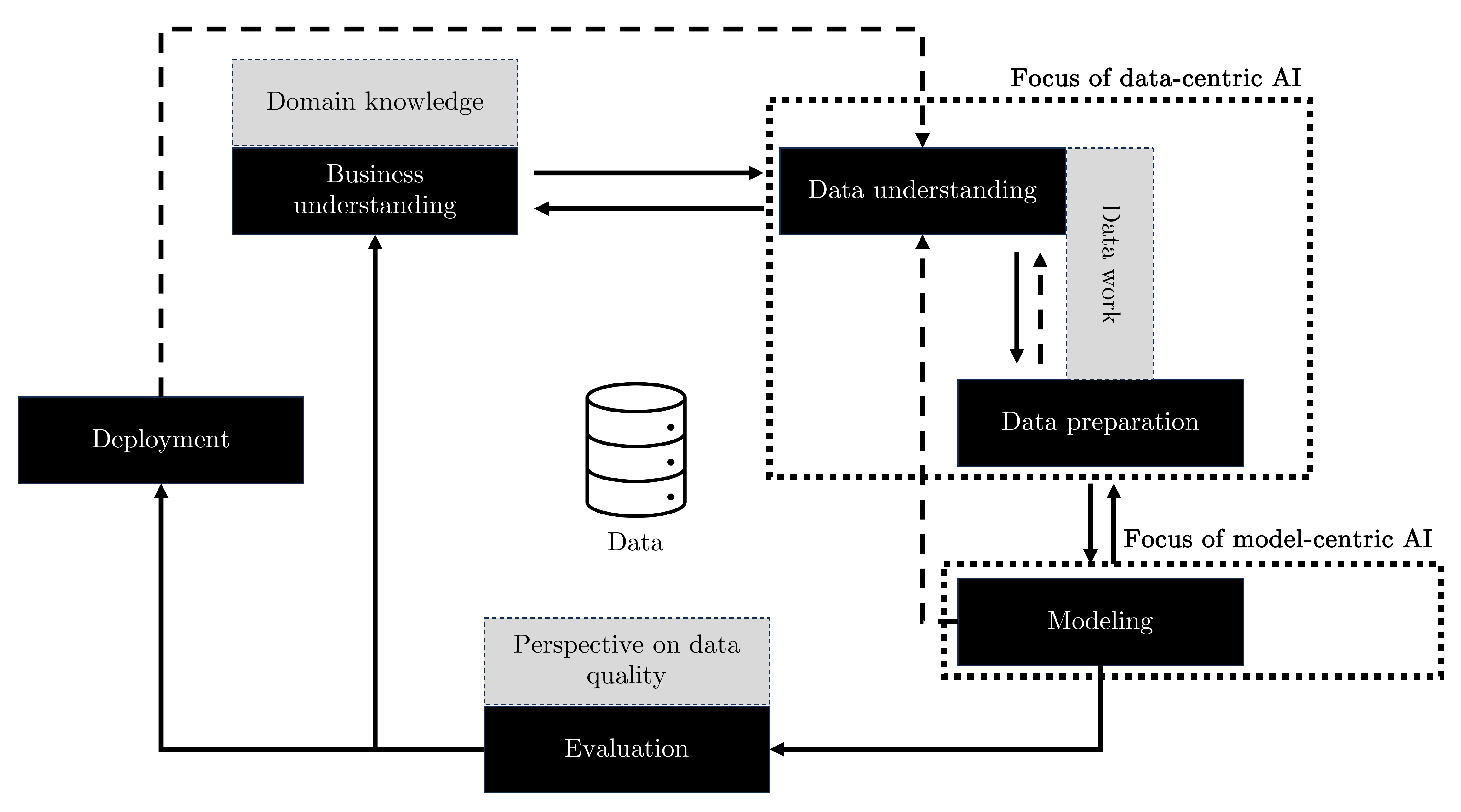}
    \caption{{{Extending the Cross Industry Standard Processes for Data Mining (CRISP-DM) based on considerations from data-centric AI}.\protect\footnotemark}}
    \label{fig:crisp-dm}
\end{figure}

{
{Hitherto, the development of AI and associated systems not only requires high-quality data but has largely benefitted from established processes, such as CRISP-DM \citep{shearer2000crisp}, which delineate and interconnect relevant stages in the development of data mining projects}. 
Within the CRISP-DM framework, model-centric AI primarily {focusses on the modeling stage. Conversely, data-centric AI  accentuates data work, including data understanding and data preparation, which requires domain knowledge.} 
{We expect three major modifications in the development of AI-based systems based on the emergence of data-centric AI, which we visualize in the context of CRISP-DM in Figure~\ref{fig:crisp-dm}}.
First, data-centric AI enforces an iterative process between understanding data and preparing data for subsequent modeling. 
In this iterative process, data versioning can help to keep track of changes in the dynamically adjusted and augmented data.
Second, during modeling, data-centric AI advocates for selecting the most appropriate model based on the data understanding and the domain knowledge. Initial tests of different methods during the modeling stage require revisiting data understanding (e.g., is data quantity sufficient for a specific method? See E1--E3). 
Third, continuous model improvement is a central aspect of data-centric AI. This acknowledges that data is dynamic and ever-changing, and AI models need to be continuously updated and refined to maintain accuracy and relevance \citep{BaierKellnerKuehl2021_1000125716}. {This requires continuous data work and an adjusting data understanding over time.}
In the past, data work processes were barely routinized or standardized. The shift towards data-centric AI underlines that companies need to actively manage the day-to-day activities of data work to standardize processes, methods, and tools. 
{BISE research is ideally suited to guide the process towards augmenting and implementing standard processes for data work and the development of AI-based systems in general including both theoretical and practical considerations.}

\footnotetext{{Black boxes and solid arrows indicate components of CRISP-DM, while the remaining components emerge from data-centric AI. Gray boxes refer to the key characteristics defined in Section~2.}}

\vspace{30pt}
\subsection{{Cross-Organizational Level}}

{Often, relevant data is scattered across various organizations, so that data sharing and interorganizational cooperation represent important considerations in data-centric AI. Previous research, though, has identified identified a range of barriers that prevent data sharing \citep{fassnacht2023barriers}.} BISE research can play a pivotal role in facilitating effective data sharing practices \citep{otto2019designing}. 
Researchers can support the design and implementation of data sharing platforms and infrastructures that enable data owners to publish and share datasets with others in a controlled and collaborative manner (E1--E3). 
In this context, implementing data sharing standards, such as data marketplaces with predefined quality standards, becomes crucial for optimizing data and generating business value. 
These platforms should address important aspects such as data versioning, data licensing, and data citation to ensure proper data governance. 
Additionally, exploring incentive mechanisms for data sharing can encourage organizations to share their data by providing appropriate rewards, such as data credits, collaboration opportunities, or shared benefits, fostering a culture of data sharing. 
Developing trust and reputation systems for data sharing is equally important \citep{fassnacht2023barriers}, as they can help assess the reliability and credibility of data sources, allowing data-centric AI projects to identify high-quality and trustworthy datasets. 
{Guidance from the BISE community to design those cross-disciplinary approaches is essential in addressing these complex questions and advancing the data sharing landscape for data-centric AI.}


{Cross-organizational usage of AI requires either data sharing or sharing locally trained models as part of federated learning \citep{Hirt2023}.} 
In recent years, research and practice have started sharing ML models and utilizing federated learning to mitigate a lack of data. However, when sharing models instead of data, ensuring high data quality for each individual model across organizational entities is challenging \citep{deng2021fair}. For example, without access to the entire set of data, it is difficult to assess the label quality (e.g., R2) or the relevance of instances (e.g., R3). This results in a need for methods and semi-automated tools to facilitate data work across organizational entities and distributed datasets.
Overall, data-centric AI across organizations requires distributed data understanding and data preparation across different organizational entities. 
{For cross-organizational data preparation, organizations need to agree on common standards for the processing of data. This also includes questions around the ownership of code pipelines for data preparation. From a technical perspective, common data processing requires aligned coding environments as well as versioning of code.}
The deployment of the resulting federated model may then have different implications for each organization's business understanding, domain knowledge, and data understanding (see Figure~\ref{fig:crisp-dm}).
The BISE community has significant experience in cross-organizational research and, therefore, is well-positioned to inform data-centric AI across organizations.
}

\section{{Conclusion}}
{Data is an indispensable component of any AI-based system. 
Data-centric AI and the corresponding focus on data work in the development of AI-based systems have significant implications for BISE researchers and practitioners. 
In this work, we {introduced data-centric AI as an emerging paradigm, contrasted data-centric AI with related concepts, and highlighted a range of existing gaps in the literature that will benefit from guidance from the BISE community.}
The paradigm of data-centric AI has the potential to significantly improve the performance of AI-based systems in research and practice making it a promising field to study for BISE research.
}

\newpage

\setstretch{2.0}
\bibliographystyle{spbasic-bise}
\bibliography{references}

\end{document}